\documentclass{article}

\usepackage{amsmath,amsfonts,bm}

\def\eqref#1{equation~\ref{#1}}

\def\1{\bm{1}}

\def\va{{\bm{a}}}

\def\vv{{\bm{v}}}

\def\mW{{\bm{W}}}

\def\mY{{\bm{Y}}}

\DeclareMathAlphabet{\mathsfit}{\encodingdefault}{\sfdefault}{m}{sl}
\SetMathAlphabet{\mathsfit}{bold}{\encodingdefault}{\sfdefault}{bx}{n}
\newcommand{\tens}[1]{\bm{\mathsfit{#1}}}

\def\tP{{\tens{P}}}

\def\tZ{{\tens{Z}}}

\newcommand{\etens}[1]{\mathsfit{#1}}

\def\etP{{\etens{P}}}

\def\etZ{{\etens{Z}}}

\DeclareMathOperator*{\argmax}{arg\,max}

\usepackage{microtype}
\usepackage{graphicx}
\usepackage{subfigure}
\usepackage{booktabs} 
\graphicspath{ {./figures/} }

\usepackage{hyperref}

\usepackage[accepted]{icml2020}

\icmltitlerunning{Subclass Distillation}

\begin{document}

\twocolumn[
\icmltitle{Subclass Distillation}



\icmlsetsymbol{equal}{*}
\begin{icmlauthorlist}
\icmlauthor{Rafael M\"uller}{to}
\icmlauthor{Simon Kornblith}{to}
\icmlauthor{Geoffrey Hinton}{to}
\end{icmlauthorlist}

\icmlaffiliation{to}{Google Brain, Toronto, Canada}

\icmlcorrespondingauthor{Rafael M\"uller}{rafaelmuller@google.com}

\icmlkeywords{Machine Learning, ICML}

\vskip 0.3in
]



\printAffiliationsAndNotice{}  

\begin{abstract}
After a large ``teacher" neural network has been trained on labeled data, the probabilities that the teacher assigns to incorrect classes reveal a lot of information
about the way in which the teacher generalizes. By training a small ``student" model to match these probabilities, it is possible to transfer most of the generalization ability of the teacher to the student, often producing a much better small model than directly training the student on the training data. The transfer works best when there are many possible classes because more is then revealed about the function learned by the teacher, but in cases where there are only a few possible classes we show that we can improve the transfer by forcing the teacher to divide each class into many subclasses that it invents during the supervised training. The student is then trained to match the subclass probabilities. For datasets where there are known, natural subclasses we demonstrate that the teacher learns similar subclasses and these improve distillation. For clickthrough datasets where the subclasses are unknown we demonstrate that subclass distillation allows the student to learn faster and better.
\end{abstract}

\section{Introduction}

The idea of compressing a teacher model into a smaller student model one by matching the predictions of the teacher was introduced by \citet{Caruana2006model}. After training the teacher, they performed the transfer on new, unlabelled data by minimizing the squared difference between the logits of the final softmax of the teacher and student models.  A related technique, called ``distillation", was introduced by \citet{hinton2015distilling}. That paper performed the transfer on the labelled training data rather than on new, unlabelled data. The student is trained to minimize a weighted sum of two different cross entropies. The first is the cross entropy with the correct answer using a standard softmax. The second is the cross entropy with the probability distribution produced by the teacher when using a temperature higher than 1 in the softmax of both models. The point of using a higher temperature is to emphasize the differences between the probabilities of wrong answers that would all be very close to zero at a temperature of 1.


There have since been some interesting theoretical developments of distillation \cite{lopez2015unifying} and it is now being widely used to produce small models that generalize well. These are needed for resource constrained applications of neural networks such as text-to-speech \citep{oord2017parallel} and mobile on-device convolutional neural networks \citep{howard2017mobilenets}.

In this work, we focus on distillation for datasets where there are only a few possible classes, resulting in limited information to be transferred (e.g.\ binary classification). We show that we can improve the transfer by forcing the teacher to divide each class into many subclasses that it invents during the supervised training. We propose an auxiliary loss that encourages each subclass to be used equally while ensuring that each prediction is ``peaky". We show experimentally that the subclasses learned have semantic meaning and help distillation. The subclasses can also be used to interpret the models predictions by clustering them in discrete bins.


\begin{figure*}[h]
\begin{center}
\includegraphics[width=0.8\textwidth]{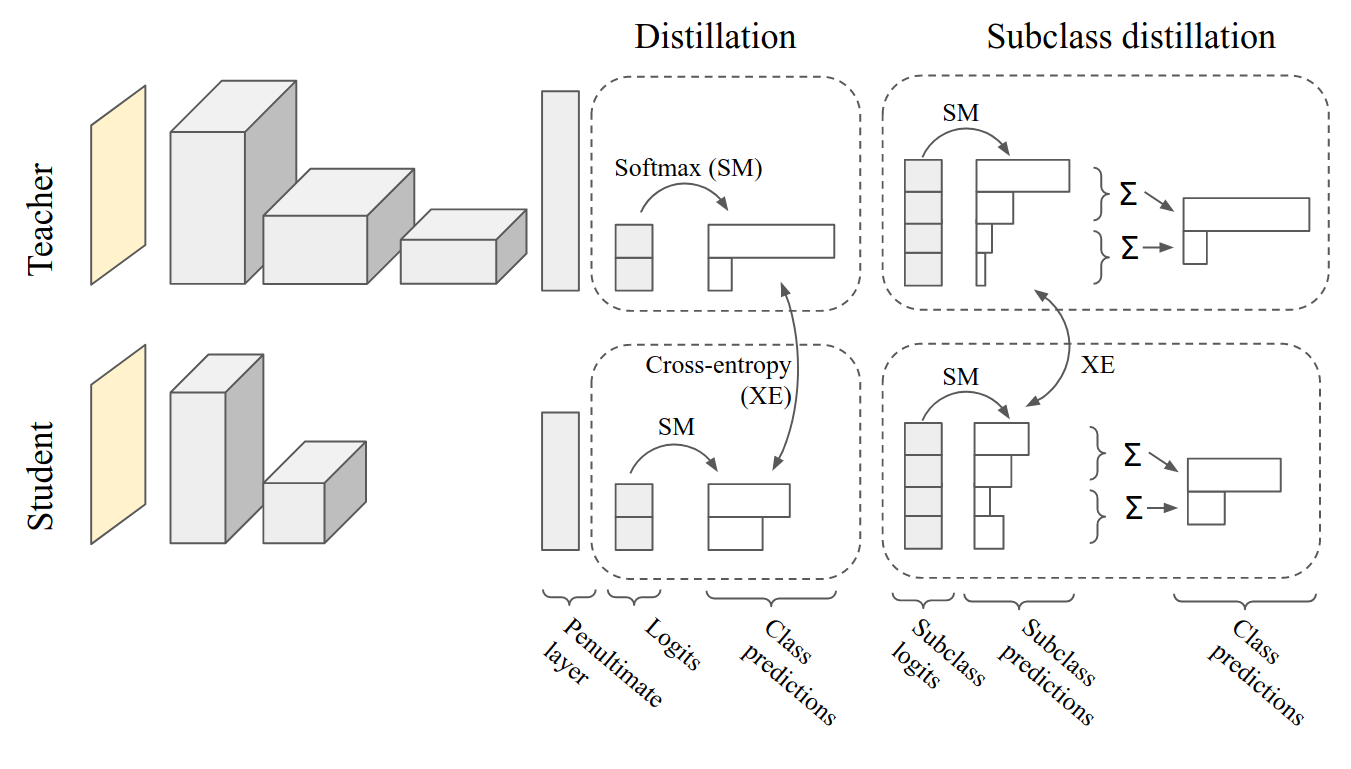}
\end{center}
\label{fig:schematic}
\vskip -1em
\caption{Comparison between distillation and subclass distillation using 2 classes and 2 subclasses per class. The teacher is usually deeper and/or wider than the student. For distillation, the student mimics (using temperature-scaled cross-entropy) the teacher's class predictions while in subclass distillation the student mimics the subclasses predictions that were invented by the teacher. The class predictions are derived by summing the subclass predictions and the only ground-truth supervision for both cases are binary class labels.}
\end{figure*}

The paper is organized as follows. We start with a description of subclass distillation and a comparison to a related method, penultimate layer distillation. First, we train models on a binary split of CIFAR-10 \citep{krizhevsky2009learning} that we call CIFAR-2x5, where we group sets of 5 classes together to create a binary classification task. We show that a teacher trained to produce subclasses is able to discover the original CIFAR-10 classes, despite receiving only binary supervision. We also show that distilling from this teacher using these learned subclasses leads to better results as compared to conventional distillation and penultimate layer distillation. We next move to the CelebA dataset \citep{liu2015faceattributes}, in which each example has 40 binary labels. We show that when predicting a single one of these binary labels, the subclasses produced by the teacher are highly correlated with the other binary labels it has never been trained on, which helps subsequent subclass distillation.

We conclude the experimental section with two additional results. First, on the Criteo click prediction dataset \citep{criteo_labs_2017}, we show that subclass distillation outperforms conventional distillation in terms of training speed. We also show that when the student does not see the full dataset, subclass distillation provides significant generalization gains. Second, using MNIST-2x5 \citep{lecun1998mnist}, we show that the student can learn to predict the binary label by learning to predict the relative subclass probabilities (intra-class), without having ever seen the binary labels or receiving class relative probabilities from the teacher.

\section{Subclass distillation}

During distillation, the amount of information that the student network receives about the generalization tendencies of the teacher network depends on the number of classes. The information provided by the hard target labels is logarithmic in the number of classes, but the information about how the teacher generalizes is linear in the number of classes provided we distill using the logits or using cross-entropy at a high temperature. This means that distillation is considerably less efficient for models with few classes.

Binary classifiers are important in many applications, and the aim of this paper is to make distillation more efficient for such models by forcing the teacher to invent $s$ subclasses for each of the $c$ classes in the dataset, as shown in Fig. \ref{fig:schematic}. The teacher computes $c \times s$ logits and puts these through a softmax to get $c \times s$ probabilities that sum to 1. The probabilities of all the subclasses of a class are then added to get the teacher's predicted probability for that class. The teacher is trained by minimizing the cross-entropy with the class probabilities:
\begin{align}
    \label{eq:lxent}
    \mathcal{L}_\text{xent} = -\frac{1}{n} \sum_{i=1}^n \sum_{j=1}^c \mY_{i, j} \log \left(\sum_{k=1}^s \etP_{i,j,k} \right)
\end{align}
where $\mY_{i,j} \in \{0, 1\}$ are the correct targets for the $j^\text{th}$ class of the $i^\text{th}$ example as by the dataset and $\etP_{i, j, k}$ is the output probability for the $k^\text{th}$ subclass of that example. Given logits $\etZ$, the output probabilities $\etP$ are computed in the usual fashion by performing a softmax operation over all logits belonging to the same example:
\begin{align}
    \etP_{i, j, k} &= \frac{\exp(\etZ_{i, j, k} / T)}{\sum_{l=1}^c\sum_{m=1}^s\exp(\etZ_{i, l, m} / T)}.
\end{align}
The temperature parameter $T$ controls the entropy of the output distribution. When training the teacher, it is set to 1. When distilling knowledge from the teacher to the student, it is often beneficial to increase the temperature.

In subclass distillation, as in conventional distillation, the student is trained to match the teacher. However, rather than use only the $c$ classes in the original dataset, the student learns to mimic the teacher's output for $c \times s$ subclasses. Like the teacher, the student produces $c \times s$ output probabilities $\tilde \tP_{i, :, :}$ for each example $i$, resulting in the subclass distillation loss:
\begin{align}
    \mathcal{L}_\text{distill} &= -T^2 \frac{1}{n} \sum_{i=1}^n \sum_{j=1}^c \sum_{k=1}^s \etP_{i, j, k} \log \left( \tilde \etP_{i, j, k} \right),
\end{align}
where we scale the loss by $T^2$ in order to keep gradient magnitudes approximately constant when changing the temperature \cite{hinton2015distilling}. Thus, with this loss, knowledge is transferred from the teacher to the student not merely through the probabilities the teacher assigns to the classes in the original dataset, but also through the probabilities assigned to the subclasses.\footnote{In conventional distillation the cross-entropy loss is $\mathcal{L}_\text{distill} = -T^2 \frac{1}{n} \sum_{i=1}^n \sum_{j=1}^c  \etP_{i, j} \log \left( \tilde \etP_{i, j} \right)$ since the teacher only produces class probabilities.} When training the student, we typically use a combination of the distillation loss $\mathcal{L}_\text{distill}$ and the standard cross-entropy loss $\mathcal{L}_\text{xent}$:
\begin{align}
    \mathcal{L}_\text{student} &= \alpha \mathcal{L}_\text{distill} + (1-\alpha)\mathcal{L}_\text{xent} 
\end{align}
where $\alpha \in [0, 1]$ controls the balance between hard and soft targets, which we call ``task balance".

\subsection{Penultimate layer distillation}
\label{sec:pld}
An alternative to subclass distillation that also incorporates more information into distillation is to distill not from the logits, but from the penultimate layer's activations (or from other layers as in \citet{romero2014fitnets}). In this case:
\begin{align}
    \mathcal{L}_\text{distill} &=  \frac{1}{n} \sum_{i=1}^n \|\va_{i}- \mW \tilde\va_{i}\|^2.
\end{align}
where $\tilde\va_{i}$ are the penultimate layer's activations of the student for the $i^{th}$ example in the minibatch, $ \va_{i}$ are the respective activations in the teacher and $\mW$ is a projection matrix to match the dimensions of teacher/student \textit{learned in the distillation phase}. Note that, the student will use its capacity to match the teacher's representations even for directions that may not be relevant for predicting the classes.

In subclass distillation, the teacher's subclass logits are a projection of the teacher's penultimate layer activations into a lower dimension which is \textit{learned during the teacher's training phase}. Therefore, the projection into subclasses can remove irrelevant information present in the penultimate layer while retaining more information compared to the ``class" logits.    

Note that \citet{hinton2015distilling} shows that minimizing the squared difference between the zero-meaned logits of the teacher and student is the limit of distillation as the temperature goes to infinity, provided that the learning rate is scaled as the squared temperature. Therefore, subclass distillation, as the temperature goes to infinity, is equivalent to penultimate layer distillation applied not on the full penultimate layer, but on a low-dimensional projection of that layer. 

\section{Auxiliary loss}
In subclass distillation, the cross-entropy loss (Eq.~\ref{eq:lxent}) constrains only the class probabilities and not the subclass probabilities. Without an additional loss encouraging the network to use  all subclasses, it may consistently assign high probability to a single subclass of each class and assign extremely low probability to the others. In this case, the subclasses would provide almost no additional signal for distillation. We thus propose an auxiliary loss that encourages the network to assign different examples to different subclasses, even when they belong to the same class. Given a minibatch of $n$ logit vectors $\vv_i = \text{vec}(\tZ_{i, :, :})$, we compute:
\begin{align}
    \mathcal{L}_\text{aux} &= -\frac{1}{n} \sum_{i=1}^n \log \frac{e^{\hat{\vv}_i^\text{T} \hat{\vv}_i/T}}{\frac{1}{n} \sum_{j=1}^n e^{\hat{\vv}_i^\text{T} \hat{\vv}_j/T}}\\
    &= \frac{1}{n} \sum_{i=1}^n \log\left(\sum_{j=1}^ne^{\hat{\vv}_i^\text{T} \hat{\vv}_j/T}\right) -\frac{1}{T} - \log(n),
\end{align}
where $\hat{\vv}_i$ is a normalized version of ${\vv}_i$ (zero-mean, unit-variance) to prevent easy solution of the minimization by making the logits large.
As above, $T$ is a temperature hyper-parameter, although its value need not correspond to the temperature used for distillation. This auxiliary loss encourages the normalized logit vector corresponding to each example to have a low dot product with other normalized logit vectors. In practice, the network accomplishes this by distributing examples across subclasses.


The total loss for the teacher is:
\begin{align}
    \mathcal{L}_\text{teacher} &= \mathcal{L}_\text{xent} + \beta\mathcal{L}_\text{aux} 
\end{align}
where $\beta$ controls the strength of the auxiliary loss. 

\section{Experimental results}
\subsection{CIFAR-10}
\begin{figure}[h]
\begin{center}
\includegraphics[width=\linewidth]{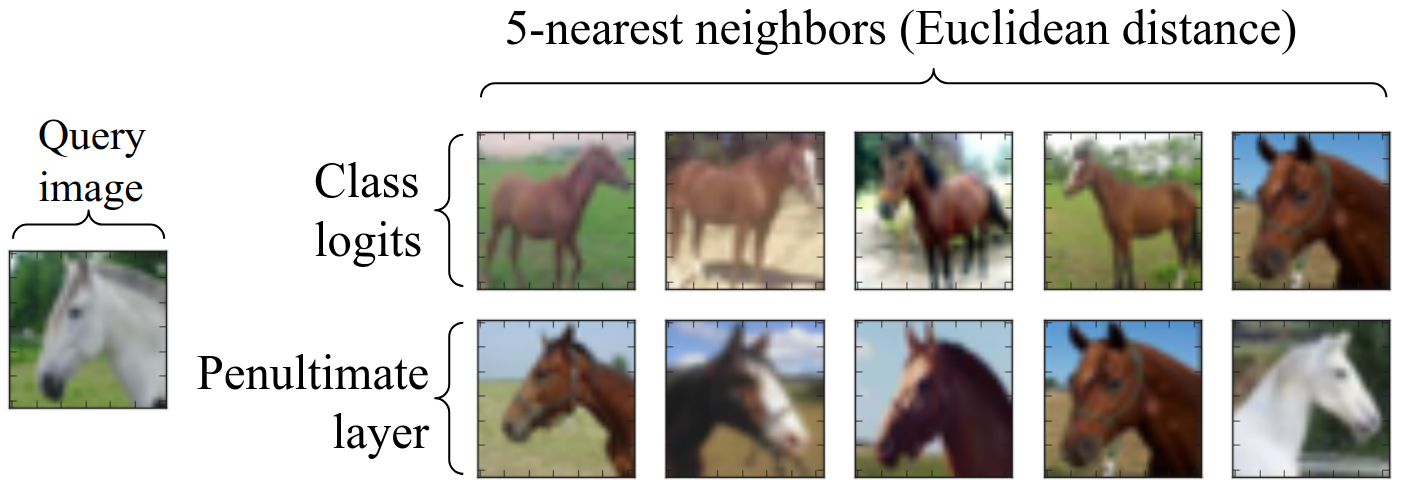}
\end{center}

\caption{Finding the nearest neighbor in a network trained on CIFAR-10. Query is a close-up on a horse's head. If the nearest neighbor is calculated in the ``class" logits layer, we find examples from the same class (horse), but the semantically similar image with a close-up head is only the $5^{th}$ nearest-neighbor. If distance is calculated in the penultimate layer, all nearest neighbors are semantically similar to the query. This shows that some semantic information is lost in the ``class" logits and distillation can benefit from using more information.}
\label{fig:nncifar10}
\end{figure}

In this section, we experimentally test the ideas presented in the previous sections.
We start by providing a visual demonstration that the hidden representations of neural networks contain semantically meaningful information that is not present in the class logits.
In Fig.\ \ref{fig:nncifar10} (top), we show the nearest neighbors using Euclidean distance in the class logits layer of a network trained on CIFAR-10 classification. We observe that the nearest neighbors are examples of the same class (horse) as we expected. However, if instead of using the logits layer, we find the nearest neighbors in the penultimate layer, we notice that not only the closest examples are from the same class, but they are also semantically similar to the query image (horse head). This is the sort of information that is present in the penultimate layer but not in the logits that we want to use to improve distillation.

Next, we move to the quantitative results. We use the CIFAR-10 dataset to construct an artificial binary classification task where we group together examples from the classes airplane, automobile, bird, cat and deer to construct the first class and dog, frog, horse, ship and truck to construct the second one. We call this task CIFAR-2x5 and by using this artificial construction we have natural semantic subclasses corresponding to the original CIFAR-10 classes. 

\subsubsection{Unsupervised subclass classification}
We train a ResNet \citep{he2016deep} network with 20 layers to be used as a teacher
(see results in Table~\ref{tab:teacher-results} and training details including hyperparameters in Appendix~\ref{app:experimental_setup}). We first train this network on CIFAR-10 as a baseline and obtain 93.5\% accuracy (averaged over 3 runs as all the results in this section). We use the same network with frozen weights to evaluate how well it does on the binary classification task and we obtain 95.6\% (+2.1\%).  If we train this network directly on the binary classification task (CIFAR-2x5), we get 94.3\%. Note that although it is evaluated on the same task, the first network is trained with 3.32 ($\log_{2}{10}$) label bits per example compared to only 1 label bit per example in the second network. This difference in the number of bits of label information explains the 1.3\% accuracy gap between them in the binary classification task and the benefit of using ``subclass" information even when the evaluation is done at the ``class" level.

\begin{table}[t]
\caption{Teacher/ResNet results over 3 runs trained on CIFAR-10 or CIFAR-2x5 and top-1 accuracy evaluation on both tasks. Additionally, we evaluate the effect of adding a subclass head and auxiliary loss on unsupervised subclass classification. For reference we include the state-of-the art result on fully unsupervised CIFAR-10 using the invariant information clustering method (IIC) in last line.}
\label{tab:teacher-results}
\vskip 0.15in
\begin{center}
\begin{small}
\begin{sc}
\begin{tabular}{lcccr}
\toprule
CIFAR- &head    &aux. loss      &acc. (2)                &acc. (10)\\
\midrule
10     &        &               &95.6$\pm$ 0.1           & 93.5 $\pm$ 0.2\\
2x5    &        &               &94.3$\pm$ 0.2           & \\
2x5    &$\surd$ &               &94.2$\pm$ 0.2           & 39.3 $\pm$ 4.0\\
2x5    &$\surd$ &$\surd$        &94.2$\pm$ 0.0           & 64.6 $\pm$ 4.8 \\
\midrule
\multicolumn{4}{l}{Unsupervised \citep{ji2018invariant}}                       &57.6 $\pm$ 5.0  \\

\bottomrule
\end{tabular}
\end{sc}
\end{small}
\end{center}
\vskip -0.1in
\end{table}

Next, we investigate how making the teacher ``invent" subclasses affects the network performance. The subclass head enables the network to output 10 logits (5 subclasses per class) which are marginalized (after softmax) over the subclasses before binary cross-entropy loss. Simply adding the head produces no improvement in binary classification despite the increase in the number of parameters in the last layer by a factor of 5. We also measure the accuracy of this network on all the 10 classes by directly taking the $\argmax$ of the subclass layer and picking the permutation that maximizes the accuracy. Although the result of 39.3\% is better than chance (20\%\footnote{Corresponding to perfect knowledge of the class and random choice of the subclass.}), we observed that since there is nothing encouraging the network to use all subclasses, they can ``die" during training. The subclass accuracy can significantly be improved by adding the auxiliary loss which increases the accuracy to 64.6\%. Note that this network has only seen binary labels, but is able to separate the classes in meaningful subclasses without extra supervision.

\begin{figure}[h]
\begin{center}
\includegraphics[width=0.45\textwidth]{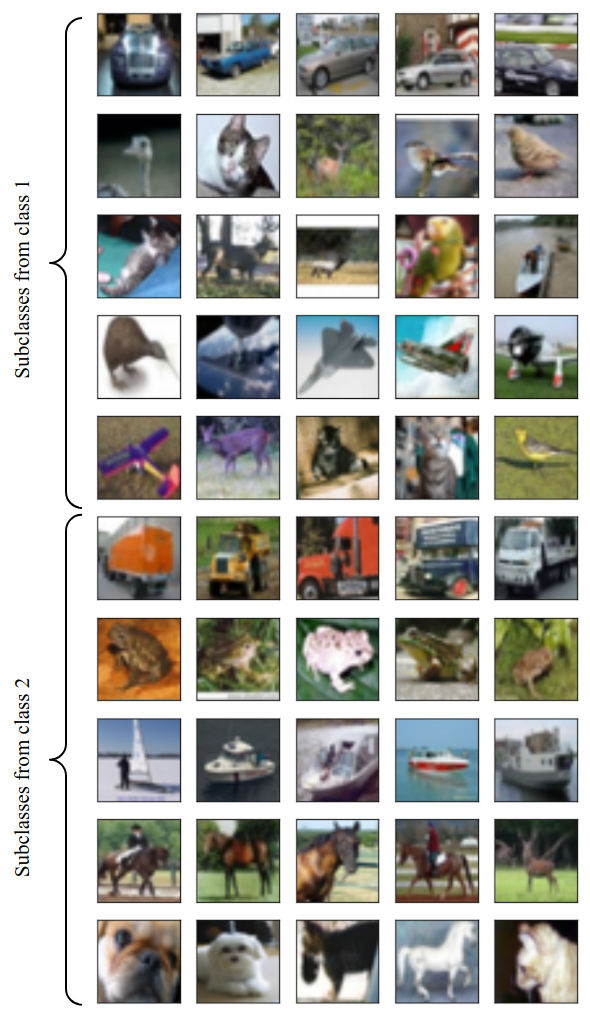}
\end{center}
\vskip -0.1in
\caption{Unsupervised subclass discovery. Examples of the validation set grouped by the subclass logit they activate most (one row per subclass). Using the validation set, we find the 1-to-1 assignment that maximizes accuracy, resulting in the following permutation: automobile, cat, bird, airplane, deer (first class), truck, frog, boat, horse and dog (second class).}
\label{fig:sub_class}
\end{figure}

Fig.\ \ref{fig:sub_class} shows how the best network out of 3 runs (70.2\%) splits a subset of examples in the validation set into subclasses.
Most errors arise in distinguishing among cats, birds and deer, while other subclasses correspond to the original dataset classes.
For comparison, the state-of-the-art \citep{ji2018invariant} on fully unsupervised classification on CIFAR-10 is 57.6\% using the invariant information clustering method (IIC). Here, we show that, with little extra supervision, (binary labels) we can outperform this result with a very simple approach.

In the analysis above, we use the accuracy on 10-class classification as a measure of how well the network separates the examples into meaningful subclasses. The idea is that this subclass information will help the student generalize better through subclass distillation. We can use a very simple model to measure how much extra label information the subclass teacher can provide. In the ideal case where the teacher perfectly learns the subclasses, it provides 1 + 2.32 label bits ($\log_{2}{2} + \log_{2}{5}$) per example, where the first bit comes from the binary class and the remaining ones from the subclass. In the case where the teacher can ``relabel" $P\times100$\% of the subclasses correctly and the remaining errors are distributed equally over the remaining 4 subclasses, the effective number of label bits is given by the $q$-ary symmetrical channel \citep{cover2012elements} and is equal to $\log_{2}{5} + P\log_{2}{P} + (1-P)\log_{2}{(1-P)/4}$. The teacher trained with binary classification + the subclass head + the auxiliary loss gets on average 67.7 $\pm$ 4.5\% subclass accuracy on the training set. This result is slightly better than results from Table \ref{tab:teacher-results} from validation set, but they are relevant for the analysis since with distillation we reuse the training set in the transferring phase. The best of the 3 runs gets 73.0\%, which results in 0.94 effective extra label bits per example given by the teacher compared to a student that only sees the binary labels. This assumes that the teacher provides noisy one-hot encoded subclass labels ("hard information") to the student, while distillation can also benefit from ``soft" information (small differences in relative probabilities) which can increase the effective number of subclass bits per example, but with the simple model our subclass teacher can already provide roughly the double amount of label information per example. 

Additionally, we would like the subclass predictions for each example to be ``peaky", resulting in probability mass concentrated mostly in a single subclass. This can be translated to having low-entropy predictions. For the network trained without the auxiliary loss the average entropy is 0.13 $\pm$ 0.02 bits while it increases to 0.42 $\pm$ 0.05 bits using the auxiliary loss, which is still far away from 3.32 bits for the uniform distribution. However, just having low-entropy predictions is not enough, since, for all examples belonging to a given dataset class, the network may assign a confident prediction to the same subclass. Therefore, we would like to ensure that after making a hard decision ($\argmax$), the distribution of subclass utilization is close to the uniform distribution (high entropy). The subclass utilization entropy is 1.87 $\pm$ 0.11 bits (without) and 3.19 $\pm$ 0.02 bits (with) the auxiliary loss. This shows that the auxiliary loss helps the subclass predictions to be confident and diverse at the same time, resulting in discovery of the original subclasses for the CIFAR-2x5 example.

\subsubsection{Subclass distillation}

In this section, we investigate how to transfer the teacher's knowledge to a low capacity student. We pick the AlexNet architecture as the student \citep{krizhevsky2012imagenet}.
Results are shown in Table \ref{tab:student-results}. We start by training the network on the two tasks without distillation, as a baseline. We observe a gap of 2.2\%  between a network trained with subclass labels (CIFAR-10) and a network without access to this extra information (CIFAR-2x5). Next, we train the student in two different situations. First we use conventional distillation.
We observe a 1.0\% accuracy gain compared to the baseline student. Then we train the same student with penultimate layer distillation and we get similar gain to conventional distillation: 1.0\% accuracy gain. 
Finally, we test subclass distillation, where we distill from a teacher that was trained to perform binary classification, but with the subclass head and auxiliary loss.
With subclass distillation, we observe a 2.3\% accuracy improvement compared to the baseline student.
The subclass distillation student can also classify the examples over 10 classes with 68.3\% accuracy which is slightly below the teacher (70.2\% which was the best of 3 runs). Note that the student trained with subclass distillation can completely recover the 2.2\% gap between the models trained with hard targets on CIFAR-10 and CIFAR-2x5 without ever seeing the ``true" subclass labels.

\begin{table}[t]
\caption{Student/AlexNet results over 3 runs. The baselines (two first rows) correspond to training the network with only the labels in the dataset. Then the distillation results correspond to training the student to match the teacher's class predictions (\textbf{D}istillation), to match the penultimate layer's activations (\textbf{P}enultimate \textbf{L}ayer \textbf{D}istillation) or the teacher's subclass predictions (\textbf{S}ub\textbf{C}lass \textbf{D}istillation).}
\label{tab:student-results}
\vskip 0.15in
\begin{center}
\begin{small}
\begin{sc}
\begin{tabular}{lccccr}
\toprule
CIFAR- &D&PL-D&SC-D&acc. (2)&acc. (10)\\
\midrule
10     &        &        &       &91.3$\pm$ 0.1           & 86.7 $\pm$ 0.2\\
\midrule
2x5    &        &        &       &89.1$\pm$ 0.2           & \\
2x5    &$\surd$ &        &       &90.1$\pm$ 0.1           & \\
2x5    &&$  \surd$       &       &90.1$\pm$ 0.1           & \\
2x5    & & &  $\surd$     &\bf{91.4}$\pm$ 0.2           & 68.3 $\pm$ 0.2 \\

\bottomrule
\end{tabular}
\end{sc}
\end{small}
\end{center}
\vskip -0.1in
\end{table}

\subsubsection{Training speed}
\begin{figure}[h]
\begin{center}
\includegraphics[width=0.5\textwidth]{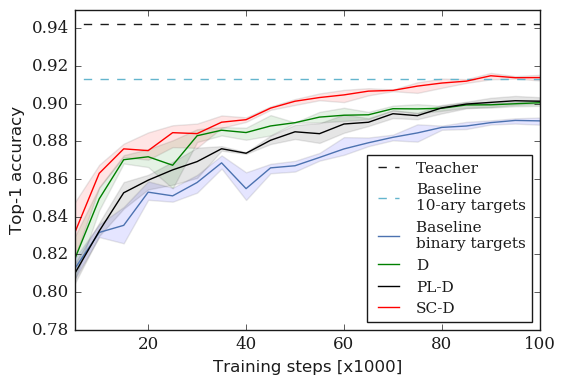}
\end{center}

\caption{CIFAR-2x5: Evolution of validation accuracy of a student (AlexNet) during training and comparison between: training only with dataset labels (baseline binary targets), distillation (D), penultimate layer distillation (PL-D) and our proposed solution, subclass distillation (SC-D). For reference, we add the performance of the teacher (ResNet-20) trained on binary labels and a student trained on 10-ary labels but evaluated on binary classification (baseline 10-ary targets).}
\label{fig:training_speed}
\end{figure}

In addition to improving performance, subclass distillation also makes training faster.
Figure \ref{fig:training_speed} shows the evolution of accuracy on the validation set through training. First we train a baseline network using only the dataset's ``hard" labels represented by the blue curve and the second row in Table \ref{tab:student-results}. We observe a large variation of performance early in training and performance increase is slow. When we train the student with conventional distillation (D), shown in green, training progresses much faster, and the final performance is better than the baseline. Since the teacher provides only a single real number per training example, there is not much information to enable the student to significantly outperform the baseline. Subclass distillation (SC-D), shown in red, addresses this issue. This results in faster training, more stable performance and higher final accuracy, matching a student trained directly on the ``true hidden" subclasses (blue dashed line). Note that both the subclass teacher and student have only seen binary labels. Finally, we show the results of penultimate layer distillation (PL-D). Although the performance is similar to distillation, training is slower, as the student tries to match the 128-dimensional teacher's activations, which may have directions that are not important for final classification.

\subsection{CelebA}

Although CIFAR-2x5 is suitable to demonstrate the subclass distillation concept and we can show significant gains in performance and training speed, the fact that the true subclass structure matches our choice of the number of subclasses makes the task easier. Therefore, we decided to test our approach on CelebA, a more realistic and challenging dataset.

\begin{figure}[h]
\begin{center}
\includegraphics[width=0.5\textwidth]{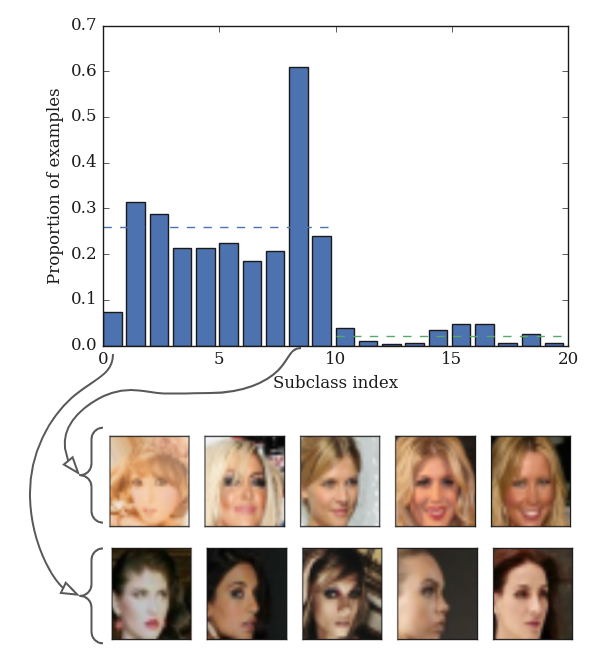}
\end{center}

\caption{CelebA: proportion of examples per subclass that have the ``blond-hair" feature. We highlight some examples of subclass ``0" and ``8",  where we observe that our teacher network splits the dataset in semantic meaningful subclasses. Dashed lines represent the class-average (female/male).}
\label{fig:celeba}
\end{figure}

CelebA comprises 202,599 images of celebrity faces, annotated with 40 binary attributes that are highly correlated and unbalanced. We pick the male/female classification task and we use 10 subclasses per class, which does not match the number of features. We obtain 1.51\% error rate using a ResNet-20 network (averaged over 3 runs). 
For some of the annotated labels, we can find a corresponding subclass that is activated by said feature. For example, in Fig. \ref{fig:celeba}, we show the proportion of examples in the validation set labeled ``blond" in each subclass, where the first 10 subclasses represent the ``female" class and the remaining the ``male" one. Dashed lines represent the average of the class (more female than male blonds in the dataset). We highlight examples that activate the first and ninth subclass and we observe that indeed the teacher has split the predictions into semantic subclasses and we speculate that this helps distillation. 




Next, we transferred knowledge from the teacher (ResNet-20) to a student (AlexNet). Results are shown in Table \ref{tab:celeba-results} in terms of error rate for the male/female prediction. The teacher achieves 1.51\% error rate while a student trained only with the hard labels achieves 2.05\%. Using conventional distillation, the error drops to 1.83\% while with subclass distillation we achieve the best performance of 1.70\%. This shows that the learned subclass factorization is useful for distillation and helps the student generalize better.

\begin{table}[t]
\caption{CelebA: results over 3 runs. 
Both the teacher (ResNet-20) and student (AlexNet) are trained with to predict the binary male/female label.
Distillation results correspond to training the student to match the teacher's class predictions (\textbf{D}istillation) or the teacher's subclass predictions (\textbf{S}ub\textbf{C}lass \textbf{D}istillation).}
\label{tab:celeba-results}
\vskip 0.15in
\begin{center}
\begin{small}
\begin{sc}
\begin{tabular}{lccccr}
\toprule
Net&D&SC-D&Error rate\\
\midrule
Teacher     &        &               &1.51$\pm$ 0.04           \\
\midrule
Student    &        &               &2.05$\pm$ 0.07           \\
Student    &$\surd$ &               &1.83$\pm$ 0.05            \\
Student    &&$  \surd$              &\bf{1.70}$\pm$ 0.12            \\

\bottomrule
\end{tabular}
\end{sc}
\end{small}
\end{center}
\vskip -0.1in
\end{table}

\subsection{Criteo}

In our CIFAR-2x5 and CelebA experiments, we ignored some of the available supervision during training time and instead used it for evaluation, in order to verify that our approach learns meaningful subclasses.
In a real-world scenario, we would use all the available information for training. Therefore, we also tested our approach on a binary dataset without a known subclass structure, the Criteo click prediction dataset \cite{criteo_labs_2017}.
This dataset consists of anonymized real-valued and categorical features. The target is a binary label indicating whether the ad was clicked.

Subclass distillation accelerates training on the Criteo dataset and leads to accuracy improvements when limited data is used for distillation. We use the large version of this dataset and we downsample the non-click examples to create a balanced dataset. The teacher is a 5-layer fully-connected network achieving 71.5\% accuracy, while the student is a 1-hidden layer network achieving 71.4\%. Note that a tiny accuracy improvement is significant in click prediction tasks since it results in large revenue increase for large user bases \citep{wang2017deep}. We then compare distillation to subclass distillation. Both achieve 71.6\% accuracy, which is better than the teacher. More important, subclass distillation again trains faster, as it provides more information about teacher generalization per example, but the dataset is so big this does not affect final performance. If we artificially reduce the amount of data that the student is trained on (10\% of the total) to exaggerate the performance difference, then we observe accuracy gains by using subclass distillation. The ability to perform distillation with limited data is attractive for large datasets such as Criteo (over 1 terabyte in size). 

\begin{figure}[h]
\begin{center}
\includegraphics[width=0.5\textwidth]{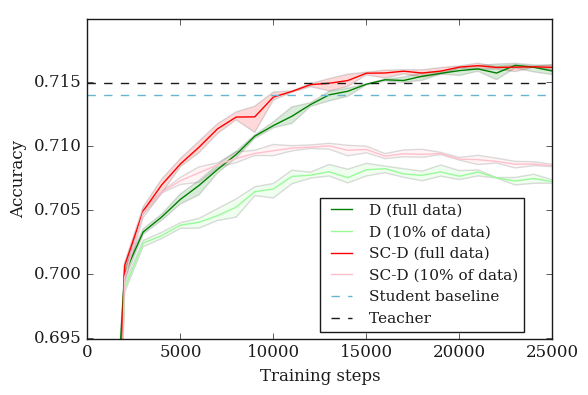}
\end{center}
\label{fig:nn}
\caption{Criteo click prediction: Evolution of validation accuracy of a student during training and comparison between: distillation (D) and subclass distillation (SC-D). When the transfer set contains all the training data, SC-D trains faster but final performance is comparable. By reducing the transfer set by a factor of 10, we exagerate the performance gap and SC-D outperforms D as it provides the student more bits per training example.}
\end{figure}
\subsection{MNIST}

\begin{figure}[h]
\begin{center}
\includegraphics[width=0.5\textwidth]{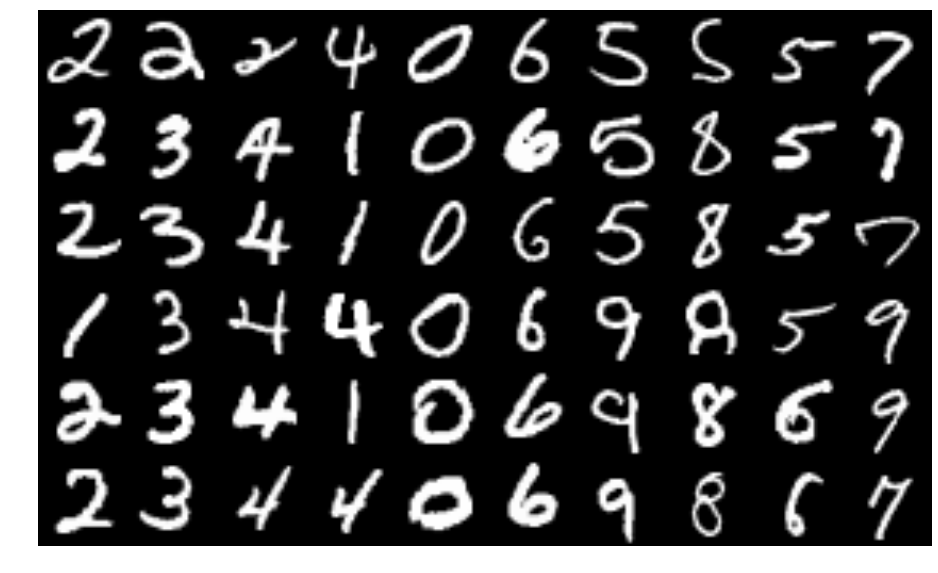}
\end{center}
\caption{MNIST unsupervised subclass discovery. Examples of the validation set and which subclass logit they activate most (one column per subclass).}
\label{fig:mnist}
\end{figure}

As our final experiment, we split the MNIST dataset into a binary classification (MNIST-2x5), by grouping digits 0 to 4 in one class and digits 5 to 9 in the other. We train a convolutional teacher to produce 10 subclasses. Fig. \ref{fig:mnist} shows how the network groups the examples into subclasses (each column represents one subclass). This network achieves 0.73\% $\pm$ 0.09 error rate in the binary classification task. A fully connected 2 hidden layer student achieves 1.57\% $\pm$ 0.06. We then distill the teacher using distillation (1.23\% $\pm$ 0.04) while subclass distillation achieves 0.93\% $\pm$ 0.06.

More interestingly, we can train the student without the hard targets by encouraging the student to mimic the intra-class relative probabilities provided by the teacher. We apply a separate softmax to each group of subclass logits to keep relative intra-class probabilities and erase relative class probability.  Then we train the student with two  cross-entropy losses over 5 subclasses, one per class. This way, the student never sees the binary label, but surprisingly learns it indirectly, obtaining 2.06\% $\pm$ 0.18 error rate. This is analogous to the experiment in Section 3 of \citet{hinton2015distilling}, where the authors omit the digit ``3" in the transfer set and the networks learned to correctly classify them just by observing the soft prediction of the digit ``3" for the remaining digits it has seen.

\section{Related work}
Several distillation methods have been proposed in the last few years \citep{dist1,dist2,dist3,dist4,dist5,dist6,dist7,dist8,dist9,dist10}. Some methods focus on teachers and students with the same architecture which can be trained sequentially \citep{furlanello2018born,xie2019self} (using unlabeled data and noisy student), \citep{bagherinezhad2018label} (using extensive data augmentation) or in parallel \citep{anil2018large} (ensemble). Other methods distill from earlier layers using $L_2$ loss \citep{romero2014fitnets, sun2019patient}. The relationship between our method and these methods is described in section \ref{sec:pld}. Recently, \citet{tian2019contrastive} proposed to distill from the penultimate layer using a contrastive loss. The relationship between our approach contrastive distillation is more vague; we use a contrastive loss during the teacher training phase to learn the subclasses while in their method it is used during distillation phase.

Our method also bears some resemblance to clustering methods. \citet{ji2018invariant} use a contrastive loss similar to our auxiliary loss (they use pairs of data augmented examples to create an anchor, whereas our loss effectively pairs the example with itself) to obtain state-of-the art results on CIFAR-10 in unsupervised and semi-supervised settings. A similar loss has been used for representation learning in \citep{hjelm2018learning, tian2019contrastive,he2019momentum, oord2018representation}. In these works, the loss is applied either in an unsupervised setting, or in a semi-supervised setting where only part of the dataset has labels. By contrast, in our case, all examples have a binary label, and we want to learn the hidden subclass labels. Moreover, these methods learn a high dimension representation of the data, whereas we learn exactly the number of subclasses with no need for a linear layer on top. An alternative method for unsupervised clustering with deep neural networks that is not based on the contrastive loss can be found in \citet{kosiorek2019stacked}, where they use capsule networks to directly learn MNIST and CIFAR-10 classes. The closest method to ours is that of \citet{krause2010discriminative}, which also uses a probabilistic classifier for clustering by optimizing for class balance and class separation, although the authors use a different loss for this purpose, and perform experiments with kernel methods rather than deep neural networks.

\section{Conclusion}

We propose subclass distillation, a distillation method where the teacher divides each class into many subclasses that it invents, and the student matches these subclass probabilities. We show that we can improve learning compared to conventional distillation and penultimate layer distillation in terms of generalization and/or training speed. We showed that with a simple auxiliary loss, our teacher divides examples of the dataset into semantically meaningful subclasses. The loss encourages the subclass predictions to be confident and diverse. 

We showed that when the underlying subclass structure is known and matches the choice of number subclasses (CIFAR-2x5 and MNIST-2x5), we can discover the original subclasses with high accuracy, and subclass distillation outperforms other distillation methods. When there is a subclass structure in the dataset which does not match the number of subclasses chosen (CelebA), our method can still discover semantic subclasses which help subclass distillation. Finally, when there is no known subclass structure (Criteo), subclasses can provide faster transfer and more bits per example when the data available is limited. We further validated that subclass distillation provides additional bits per example by showing on MNIST that we can learn to predict the binary label without any binary supervision, just by mimicking the (intra-class) teacher subclass relative probabilities.  

\bibliography{example_paper}

\begin{thebibliography}{36}
\providecommand{\natexlab}[1]{#1}
\providecommand{\url}[1]{\texttt{#1}}
\expandafter\ifx\csname urlstyle\endcsname\relax
  \providecommand{\doi}[1]{doi: #1}\else
  \providecommand{\doi}{doi: \begingroup \urlstyle{rm}\Url}\fi

\bibitem[Ahn et~al.(2019)Ahn, Hu, Damianou, Lawrence, and Dai]{dist4}
Ahn, S., Hu, S.~X., Damianou, A., Lawrence, N.~D., and Dai, Z.
\newblock Variational information distillation for knowledge transfer.
\newblock In \emph{Proceedings of the IEEE Conference on Computer Vision and
  Pattern Recognition}, pp.\  9163--9171, 2019.

\bibitem[Anil et~al.(2018)Anil, Pereyra, Passos, Ormandi, Dahl, and
  Hinton]{anil2018large}
Anil, R., Pereyra, G., Passos, A., Ormandi, R., Dahl, G.~E., and Hinton, G.~E.
\newblock Large scale distributed neural network training through online
  distillation.
\newblock \emph{arXiv preprint arXiv:1804.03235}, 2018.

\bibitem[Bagherinezhad et~al.(2018)Bagherinezhad, Horton, Rastegari, and
  Farhadi]{bagherinezhad2018label}
Bagherinezhad, H., Horton, M., Rastegari, M., and Farhadi, A.
\newblock Label refinery: Improving imagenet classification through label
  progression.
\newblock \emph{arXiv preprint arXiv:1805.02641}, 2018.

\bibitem[Bucila et~al.(2006)Bucila, Caruana, and
  Niculescu-Mizil]{Caruana2006model}
Bucila, C., Caruana, R., and Niculescu-Mizil, A.
\newblock Model compression.
\newblock In \emph{Proceedings of the 12th ACM SIGKDD international conference
  on Knowledge discovery and data mining}, pp.\  535--541. ACM, 2006.

\bibitem[Cover \& Thomas(2012)Cover and Thomas]{cover2012elements}
Cover, T.~M. and Thomas, J.~A.
\newblock \emph{Elements of information theory}.
\newblock John Wiley \& Sons, 2012.

\bibitem[CriteoLabs(2017)]{criteo_labs_2017}
CriteoLabs.
\newblock Download terabyte click logs, Jan 2017.
\newblock URL
  \url{http://labs.criteo.com/2013/12/download-terabyte-click-logs/}.

\bibitem[Furlanello et~al.(2018)Furlanello, Lipton, Tschannen, Itti, and
  Anandkumar]{furlanello2018born}
Furlanello, T., Lipton, Z., Tschannen, M., Itti, L., and Anandkumar, A.
\newblock Born again neural networks.
\newblock In \emph{International Conference on Machine Learning}, pp.\
  1607--1616, 2018.

\bibitem[He et~al.(2016)He, Zhang, Ren, and Sun]{he2016deep}
He, K., Zhang, X., Ren, S., and Sun, J.
\newblock Deep residual learning for image recognition.
\newblock In \emph{Proceedings of the IEEE conference on computer vision and
  pattern recognition}, pp.\  770--778, 2016.

\bibitem[He et~al.(2019)He, Fan, Wu, Xie, and Girshick]{he2019momentum}
He, K., Fan, H., Wu, Y., Xie, S., and Girshick, R.
\newblock Momentum contrast for unsupervised visual representation learning.
\newblock \emph{arXiv preprint arXiv:1911.05722}, 2019.

\bibitem[Heo et~al.(2018)Heo, Lee, Yun, and Choi]{dist7}
Heo, B., Lee, M., Yun, S., and Choi, J.~Y.
\newblock Knowledge transfer via distillation of activation boundaries formed
  by hidden neurons, 2018.

\bibitem[Hinton et~al.(2014)Hinton, Vinyals, and Dean]{hinton2015distilling}
Hinton, G., Vinyals, O., and Dean, J.
\newblock Distilling the knowledge in a neural network.
\newblock \emph{NIPS 2014 Deep Learning Workshop}, 2014.

\bibitem[Hjelm et~al.(2018)Hjelm, Fedorov, Lavoie-Marchildon, Grewal, Bachman,
  Trischler, and Bengio]{hjelm2018learning}
Hjelm, R.~D., Fedorov, A., Lavoie-Marchildon, S., Grewal, K., Bachman, P.,
  Trischler, A., and Bengio, Y.
\newblock Learning deep representations by mutual information estimation and
  maximization.
\newblock \emph{arXiv preprint arXiv:1808.06670}, 2018.

\bibitem[Howard et~al.(2017)Howard, Zhu, Chen, Kalenichenko, Wang, Weyand,
  Andreetto, and Adam]{howard2017mobilenets}
Howard, A.~G., Zhu, M., Chen, B., Kalenichenko, D., Wang, W., Weyand, T.,
  Andreetto, M., and Adam, H.
\newblock {MobileNets}: Efficient convolutional neural networks for mobile
  vision applications.
\newblock \emph{arXiv preprint arXiv:1704.04861}, 2017.

\bibitem[Huang \& Wang(2017)Huang and Wang]{dist10}
Huang, Z. and Wang, N.
\newblock Like what you like: Knowledge distill via neuron selectivity
  transfer, 2017.

\bibitem[Ji et~al.(2018)Ji, Henriques, and Vedaldi]{ji2018invariant}
Ji, X., Henriques, J.~F., and Vedaldi, A.
\newblock Invariant information distillation for unsupervised image
  segmentation and clustering.
\newblock \emph{arXiv preprint arXiv:1807.06653}, 2018.

\bibitem[Kim et~al.(2018)Kim, Park, and Kwak]{dist8}
Kim, J., Park, S., and Kwak, N.
\newblock Paraphrasing complex network: Network compression via factor
  transfer.
\newblock In \emph{Advances in Neural Information Processing Systems}, pp.\
  2760--2769, 2018.

\bibitem[Kosiorek et~al.(2019)Kosiorek, Sabour, Teh, and
  Hinton]{kosiorek2019stacked}
Kosiorek, A.~R., Sabour, S., Teh, Y.~W., and Hinton, G.~E.
\newblock Stacked capsule autoencoders.
\newblock \emph{arXiv preprint arXiv:1906.06818}, 2019.

\bibitem[Krause et~al.(2010)Krause, Perona, and
  Gomes]{krause2010discriminative}
Krause, A., Perona, P., and Gomes, R.~G.
\newblock Discriminative clustering by regularized information maximization.
\newblock In \emph{Advances in neural information processing systems}, pp.\
  775--783, 2010.

\bibitem[Krizhevsky(2009)]{krizhevsky2009learning}
Krizhevsky, A.
\newblock Learning multiple layers of features from tiny images.
\newblock 2009.

\bibitem[Krizhevsky et~al.(2012)Krizhevsky, Sutskever, and
  Hinton]{krizhevsky2012imagenet}
Krizhevsky, A., Sutskever, I., and Hinton, G.~E.
\newblock Imagenet classification with deep convolutional neural networks.
\newblock In \emph{Advances in neural information processing systems}, pp.\
  1097--1105, 2012.

\bibitem[LeCun et~al.(1998)LeCun, Cortes, and Burges]{lecun1998mnist}
LeCun, Y., Cortes, C., and Burges, C.~J.
\newblock The mnist database of handwritten digits, 1998.
\newblock \emph{URL http://yann. lecun. com/exdb/mnist}, 10:\penalty0 34, 1998.

\bibitem[Liu et~al.(2015)Liu, Luo, Wang, and Tang]{liu2015faceattributes}
Liu, Z., Luo, P., Wang, X., and Tang, X.
\newblock Deep learning face attributes in the wild.
\newblock In \emph{Proceedings of International Conference on Computer Vision
  (ICCV)}, December 2015.

\bibitem[Lopez-Paz et~al.(2016)Lopez-Paz, Bottou, Sch{\"o}lkopf, and
  Vapnik]{lopez2015unifying}
Lopez-Paz, D., Bottou, L., Sch{\"o}lkopf, B., and Vapnik, V.
\newblock Unifying distillation and privileged information.
\newblock \emph{International Conference on Learning Representations}, 2016.

\bibitem[Oord et~al.(2018{\natexlab{a}})Oord, Li, Babuschkin, Simonyan,
  Vinyals, Kavukcuoglu, Driessche, Lockhart, Cobo, Stimberg,
  et~al.]{oord2017parallel}
Oord, A. v.~d., Li, Y., Babuschkin, I., Simonyan, K., Vinyals, O., Kavukcuoglu,
  K., Driessche, G., Lockhart, E., Cobo, L., Stimberg, F., et~al.
\newblock Parallel {WaveNet}: Fast high-fidelity speech synthesis.
\newblock In \emph{International Conference on Machine Learning}, pp.\
  3915--3923, 2018{\natexlab{a}}.

\bibitem[Oord et~al.(2018{\natexlab{b}})Oord, Li, and
  Vinyals]{oord2018representation}
Oord, A. v.~d., Li, Y., and Vinyals, O.
\newblock Representation learning with contrastive predictive coding.
\newblock \emph{arXiv preprint arXiv:1807.03748}, 2018{\natexlab{b}}.

\bibitem[Park et~al.(2019)Park, Kim, Lu, and Cho]{dist5}
Park, W., Kim, D., Lu, Y., and Cho, M.
\newblock Relational knowledge distillation.
\newblock In \emph{Proceedings of the IEEE Conference on Computer Vision and
  Pattern Recognition}, pp.\  3967--3976, 2019.

\bibitem[Passalis \& Tefas(2018)Passalis and Tefas]{dist6}
Passalis, N. and Tefas, A.
\newblock Learning deep representations with probabilistic knowledge transfer.
\newblock In \emph{Proceedings of the European Conference on Computer Vision
  (ECCV)}, pp.\  268--284, 2018.

\bibitem[Peng et~al.(2019)Peng, Jin, Liu, Li, Wu, Liu, Zhou, and Zhang]{dist3}
Peng, B., Jin, X., Liu, J., Li, D., Wu, Y., Liu, Y., Zhou, S., and Zhang, Z.
\newblock Correlation congruence for knowledge distillation.
\newblock In \emph{Proceedings of the IEEE International Conference on Computer
  Vision}, pp.\  5007--5016, 2019.

\bibitem[Romero et~al.(2014)Romero, Ballas, Kahou, Chassang, Gatta, and
  Bengio]{romero2014fitnets}
Romero, A., Ballas, N., Kahou, S.~E., Chassang, A., Gatta, C., and Bengio, Y.
\newblock Fitnets: Hints for thin deep nets.
\newblock \emph{arXiv preprint arXiv:1412.6550}, 2014.

\bibitem[Sun et~al.(2019)Sun, Cheng, Gan, and Liu]{sun2019patient}
Sun, S., Cheng, Y., Gan, Z., and Liu, J.
\newblock Patient knowledge distillation for bert model compression.
\newblock In \emph{Proceedings of the 2019 Conference on Empirical Methods in
  Natural Language Processing and the 9th International Joint Conference on
  Natural Language Processing (EMNLP-IJCNLP)}, pp.\  4314--4323, 2019.

\bibitem[Tian et~al.(2019)Tian, Krishnan, and Isola]{tian2019contrastive}
Tian, Y., Krishnan, D., and Isola, P.
\newblock Contrastive representation distillation.
\newblock \emph{arXiv preprint arXiv:1910.10699}, 2019.

\bibitem[Tung \& Mori(2019)Tung and Mori]{dist2}
Tung, F. and Mori, G.
\newblock Similarity-preserving knowledge distillation.
\newblock In \emph{Proceedings of the IEEE International Conference on Computer
  Vision}, pp.\  1365--1374, 2019.

\bibitem[Wang et~al.(2017)Wang, Fu, Fu, and Wang]{wang2017deep}
Wang, R., Fu, B., Fu, G., and Wang, M.
\newblock Deep \& cross network for ad click predictions.
\newblock In \emph{Proceedings of the ADKDD'17}, pp.\  1--7. 2017.

\bibitem[Xie et~al.(2019)Xie, Hovy, Luong, and Le]{xie2019self}
Xie, Q., Hovy, E., Luong, M.-T., and Le, Q.~V.
\newblock Self-training with noisy student improves imagenet classification.
\newblock \emph{arXiv preprint arXiv:1911.04252}, 2019.

\bibitem[Yim et~al.(2017)Yim, Joo, Bae, and Kim]{dist9}
Yim, J., Joo, D., Bae, J., and Kim, J.
\newblock A gift from knowledge distillation: Fast optimization, network
  minimization and transfer learning.
\newblock In \emph{Proceedings of the IEEE Conference on Computer Vision and
  Pattern Recognition}, pp.\  4133--4141, 2017.

\bibitem[Zagoruyko \& Komodakis(2016)Zagoruyko and Komodakis]{dist1}
Zagoruyko, S. and Komodakis, N.
\newblock Paying more attention to attention: Improving the performance of
  convolutional neural networks via attention transfer.
\newblock \emph{arXiv preprint arXiv:1612.03928}, 2016.

\end{thebibliography}
\bibliographystyle{icml2020}

\appendix
\clearpage
\twocolumn

\section{Experimental setup}
\label{app:experimental_setup}
\paragraph{CIFAR-2x5}
The teacher is a ResNet-20 trained for 64k steps using a minibatch size of 128. We used a cosine learning rate schedule that drops to 0 at the end of training and Nesterov momentum (0.9). The starting learning rate was 0.1, weight decay 0.0003, temperature in the auxiliary loss 2.0. For the student, we switch the architecture to AlexNet, and increased the number of training steps to 100k. For the baseline student trained from scratch without distillation, we performed a grid search over weight decay (0.0001, 0.0003, 0.001, \textbf{0.003}, 0.01, 0.03) and learning rate (0.03, 0.1, \textbf{0.3}, 1.0). For every point, we averaged the validation accuracy over 3 runs. Values achieving the maximum accuracy are highlighted in bold. We use these same basic hyperparameters for all distillation results, but distillation adds additional hyperparameters, which we again tune by grid search. For conventional distillation, we sweep temperature $T$ (1.0, 2.0, 4.0, 8.0, 16.0, \textbf{32.0}, 64.0) and task balance $\alpha$ (0.0, 0.25, 0.5, \textbf{0.75}). For subclass distillation, we also sweep temperature (1.0, 2.0, 4.0, \textbf{8.0}, 16.0) and task balance (0.0, 0.25, 0.5, \textbf{0.75}). For penultimate layer distillation we sweep the weight we give to the $L_2$ distillation loss (0.3, 1.0, 3.0, \textbf{10.0}, 30.0).

\paragraph{CelebA}
For CelebA experiments, we follow a similar procedure to the CIFAR-2x5 experiments. We optimize the ResNet-20 teacher with 10 subclasses per class, picking values of weight decay, learning rate and auxiliary loss and temperature (0.001, 0.1, 2.0). We optimize the AlexNet student trained without distillation by performing grid search over weight decay (0.00003, 0.0001, \textbf{0.0003}, 0.001, 0.003) and learning rate (0.003, 0.01, \textbf{0.03}, 0.1). We pick these values for the distillation results and tune temperature and task balance. For conventional distillation, temperature (1.0, 2.0, \textbf{4.0}, 8.0, 16.0) and task balance (0.0, 0.25, 0.5, \textbf{0.75}). For subclass distillation, temperature (1.0, 2.0, 4.0, 8.0, \textbf{16.0}) and task balance (0.0, 0.25, \textbf{0.5}, 0.75).

\paragraph{Criteo}
The teacher is a 4 layer fully connected network with ReLU nonlinearity and the following number of neurons per layer: 2048, 1024, 512, 256. We have 13 integer-valued features and 26 categorical which are embedded with dimension 32 after being hashed to 1e6 buckets. The teacher has a total number of 10 subclasses, the auxiliary loss has temperature 2.0 and is multiplied by 0.1 before being added to the hard targets cross-entropy. The student has a single hidden layer of size 256. Both are trained using minibatch size of 8192, momentum of 0.9, learning rate starting at 0 and increasing quadratically up 0.1 at 10k steps then staying constant. Teacher and student baseline results represent early stopping at 27k steps. For the distillation results we use task balance of 0.5 and temperature of 2.0.

\paragraph{MNIST}
The teacher is a deep convolutional network with ReLU nonlinearity and the following layers in sequence: convolutional with 32 output channels and kernel size of 3x3, max-pooling with kernel size of 2x2, convolutional with 64 output channels and kernel size of 2x2, max-pooling with kernel size of 2x2 and strides of 2, dropout with rate 0.5, fully connected with 128 output neurons and dropout with rate of 0.5. The teacher uses a total of 10 subclasses and auxiliary loss temperature of 1.0. The student is a 2 hidden layer fully connected network (784 neurons per layer) and relu activation. For conventional distillation and subclass distillation, the temperature is 4.0 and task balance 0.5. Every network is trained for 12 epochs using a batch size of 256 and Adam optimizer with following parameters: learning rate of 0.1, $\beta_1$ of 0.9, $\beta_2$ of 0.999 and $\epsilon$ of $10^{-7}$.

%



\end{document}